\documentclass{article}
\usepackage{ijcai22}

\usepackage{color,xcolor}
\usepackage{times}
\usepackage{soul}
\usepackage{url}
\usepackage[hidelinks]{hyperref}
\usepackage[utf8]{inputenc}
\usepackage[small]{caption}
\usepackage{graphicx}
\usepackage{amsmath}
\usepackage{amsthm}
\usepackage{booktabs}
\usepackage{algorithm}
\usepackage{algorithmic}
\usepackage{multirow}
\usepackage{setspace}
\usepackage{amssymb}
\newtheorem{theorem}{Theorem}
\newcommand{\tabincell}[2]{\begin{tabular}{@{}#1@{}}#2\end{tabular}}
\title{Function-words Adaptively Enhanced Attention Networks\\ for Few-Shot Inverse Relation Classification}

\author{
    Chunliu Dou,
    Shaojuan Wu,
    Xiaowang Zhang,
    Zhiyong Feng,
    Kewen Wang
    \affiliations 
    College of Intelligence and Computing, Tianjin University, Tianjin, China, 300350 
    \affiliations
    Tianjin University-Aishu Data Intelligence Joint Laboratory, Tianjin, China
    \emails
    \{dou123321, shaojuanwu\}@tju.edu.cn
}

\begin{document}

\maketitle

\begin{abstract}
  The relation classification is to identify semantic relations between two entities in a given text. While existing models perform well for classifying inverse relations with large datasets, their performance is significantly reduced for few-shot learning. In this paper, we propose a function words adaptively enhanced attention framework (FAEA) for few-shot inverse relation classification, in which a hybrid attention model is designed to attend class-related function words based on meta-learning. 
  As the involvement of function words bring in significant intra-class redundancy, an adaptive message passing mechanism is introduced to capture and transfer inter-class differences
  %and intra-class commonalities between instances. 
  We mathematically analyze the negative impact of function words from dot-product measurement, which explains why message passing mechanism effectively reduces the impact. Our experimental results show that FAEA outperforms strong baselines, especially the inverse relation accuracy is improved by 14.33\% under 1-shot setting in FewRel1.0.
\end{abstract}

\section{Introduction}
\begin{figure}[h]  
\centering
  \includegraphics[width=3.5in]{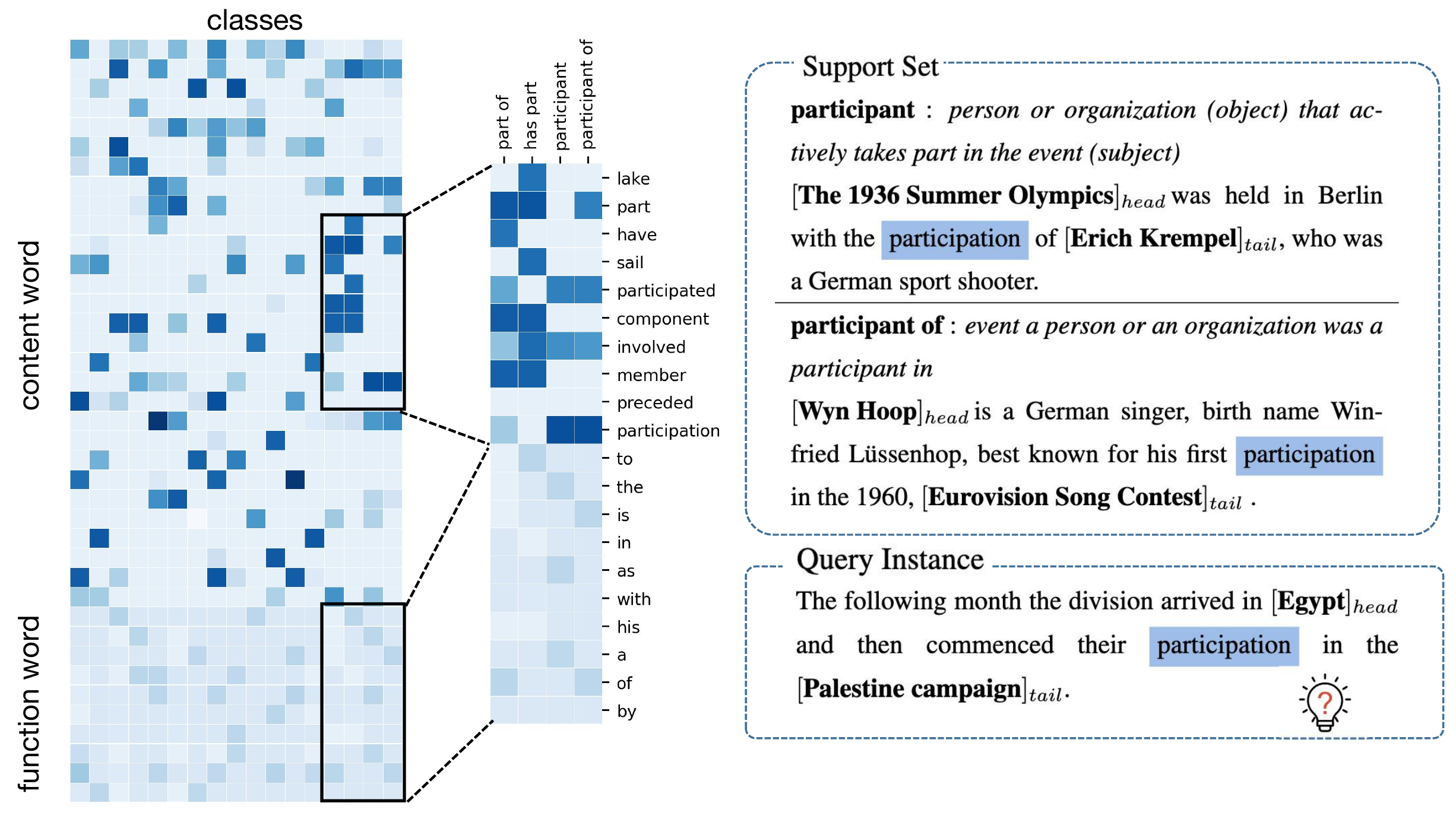}
  \caption{The left figure shows the words attention visualization of TD-PROTO, where a darker unit indicates a higher value. We observe content words region is more deeper than function words. The right is an example of RSIRC. This is a 2-way-1-shot setup-each task involves two relations, and each relation has one support instance. $[\cdot]_{\text {head}}$ and $[\cdot]_{\text {tail}}$  indicate head and tail entities respectively. }
\label{fig1}
\end{figure} 
Relation classification (RC) aims to classify the relation between two given entities based on their related context. Specifically, given a sentence in a natural language, a set of relation names, and two entities, we want to determine the correct relation between these two entities. RC is useful for many natural language processing applications, such as information retrieval \cite{bao2019few}, question answering \cite{Kaixin2021leveragin}. Most existing approaches to RC are based on supervised learning and the datasets used in training heavily depend on human-annotated data, which limit their performance on classifying the relations with insufficient instances. Therefore, making the RC models capable of handling relations with few training instances becomes a crucial challenge in AI. Inspired by the success of few-shot learning methods in the computer vision community \cite{wu2021universal,yang2022sega}, Han \emph{et al.} \cite{han2018fewrel} first investigated the problem of few-shot relations classification (FSRC) and proposed the dataset FewRel1.0 for evaluating the performance of FSRC models. Since then, several other FSRC models have been reported in the literature and they demonstrate remarkable performance on FewRel1.0. \cite{gao2019hybrid,qu2020few,yang2021entity}.

However, our experiments show that the performance of existing models for FSRC is significantly reduced when one relation is the inverse of another relation in the given set of relations.
%on few-shot distinct relations classification \cite{han2020more}, current FSRC methods are difficult to handle inverse relations as shown in Table \ref{tab3}. 
Figure \ref{fig1} shows a few-shot inverse relations classification (FSIRC) task where the set of relations contain two relations `\emph{participant}' and `\emph{participant of}'. From the two support instances, we can see that the relation `\emph{participant}' is the inverse relation of `\emph{participant of}'. We note that `\emph{participant}' and `\emph{participant of}' have the same content word `\emph{participation}' but different function words `\emph{of}' and `\emph{his}'. 
Existing models  \cite{sun2019hierarchical,bao2019few} for FSRC focus on characterising the differences between content words but ignore the differences of function words. As a result, these models do not perform well when one relation is the inverse of another relation. 

In practical applications, it is often the case that one relation is the inverse of another relation. For example, we found 21.25\% of relations in FewRel1.0 dataset are inverse. However, it is useful but challenging to classify relations with the presence of inverse relations in few-shot scenarios, as the lack of sufficient samples makes it hard to determine the important distribution of class-related function words.

\begin{figure*}
\centering
\includegraphics[scale=.4]{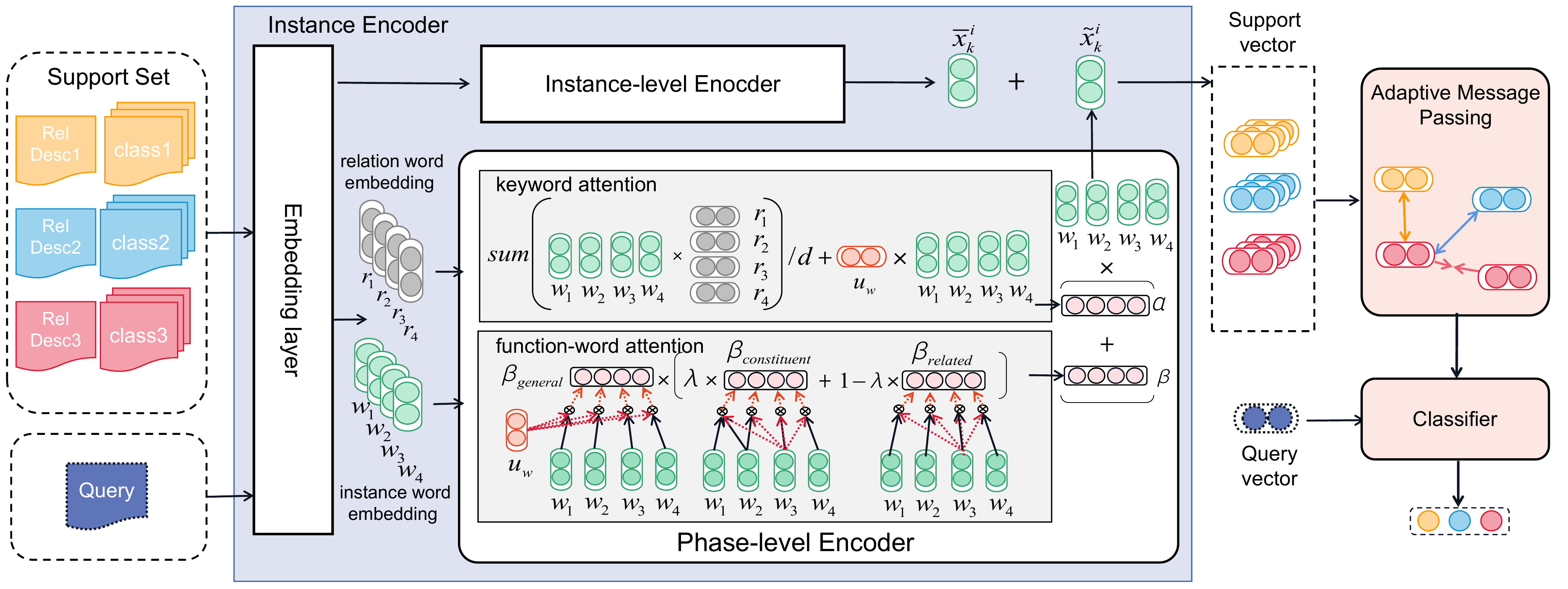}
\caption{The overall framework of FAEA. The input of Instance Encoder is an instance with corresponding relation description. }
\label{img2}
\end{figure*}

In order to address this issue, we propose a new approach to the problem of FSIRC, called \emph{Function-words Adaptively Enhanced Attention Networks (FAEA)}. As shown in Figure \ref{img2}, in the instance encoder, besides considering the importance of keywords, we use a hybrid attention to capture function words. Specifically, the class-general attention mechanism learns general function-words importance distribution. As function words appearing in the same phase with keywords are more likely to be informative \cite{zhang2020target}, we design a class-specific attention by strengthening function-words importance adjacent to keywords.
From our experience, in some cases, function-words far from keywords are also important.
For this reason, we introduce semantic-related attention for computing the direct semantic relevance between function-words and keywords. However, the introduction of function words may increase the intra-class differences. So we present a message passing mechanism to capture and transfer inter-class differences and intra-class commonalities between instances. But when inter-class differences are large, they will bring in noises and thus useful relation semantics can be lost. To avoid this issue, we adaptively control proportion of transferred inter-class message. Our experiments show that FAEA significantly outperforms major baseline models for FSIRC.

In a nutshell, our contributions are listed as follows:
\begin{itemize}
    \item We present FAEA that learns to capture class-related function words. This is achieved by introducing an adaptive message passing mechanism to obtain phase-level local features and thus discriminative representations.
    \item We mathematically show that the involvement of function words will increase intra-class differences from dot-product measurement and the designed message passing mechanism effectively reduces the redundancy.
   \item We conduct experiments on two benchmarks and show that our model significantly outperforms the baselines, especially for FSIRC task. Ablation experiments demonstrate the effectiveness of the proposed modules.  
\end{itemize}

\section{Related Work}
%\paragraph{Few shot relation classification}
Few shot relation classification aims to predict novel relations by exploring a few labeled instances. Existing methods can be mainly divided into two categories: Gradient-based models and metric-based models. A gradient-based model \cite{finn2017model,qu2020few} can rapidly adapt the model to a given task via a small number of gradient update steps. MAML\cite{finn2017model} is a representative model, learning a suitable initialization of model parameters by learning from base classes, and transferring these parameters to novel classes in a few gradient steps. Metric-based models \cite{snell2017prototypical,gao2019hybrid,ye2019multi,wen2021enhanced} leverage similarity information between samples to identify novel classes with few samples. As a representative model, Prototypical Network \cite{snell2017prototypical} takes the mean of support samples. Some models\cite{ye2019multi,wen2021enhanced} add attention mechanisms to enhance the prototypical network for highlighting crucial instances and features, but they ignore the intra-sentence difference information. Some models \cite{sun2019hierarchical,bao2019few,yang2021entity} capture local content words to obtain fine-grained information and ignore function words importance. However, inverse relations of FSRC has not been effectively handled by current models. In this work, we focus on inverse relations and propose a hybrid function words attention method to model subtle variations across inverse relations.

%\paragraph{Graph neural networks for few-short learning}
%Graph neural networks (GNNs) are employed to refine the node representations by aggregating and transforming neighboring nodes recursively. Recent approaches \cite{garcia2017few,kim2019edge} exploit GNNs in few-shot learning tasks to express complex interactions among instances. But most methods only aggregate low-frequency signals from neighbors, resulting in retaining the commonality of connected node features while ignoring transferring the difference, according to \cite{bo2021beyond}. However, intra-class commonality is scarcer than inter-class differences in few-shot learning. In this work, we design a message passing model that adaptively transfers inter-class information between instances to make representations more discriminative.

\section{Our Method}
\subsection{Problem Statement}
FSRC is defined as a task to predict the relation $y$ between the entity pair $(h, t)$ mentioned in a query instance $x^{q}$ (i.e., a sentence containing the entities $h$ and $t$), given a support set $\mathcal{S}$ and a relation set $\mathcal{R}$,
$\mathcal{S}=\left\{\left(x_{k}^{i}, h_{k}^{i}, t_{k}^{i}, r^{i}, y^{i}\right) ; i=1, \ldots, N, k=1, \ldots, K\right\}$ and $\mathcal{R}=\left\{y^{1}, y^{2}, \ldots, y^{N}\right\}$,
where $\left(x_{k}^{i}, h_{k}^{i}, t_{k}^{i}, r^{i}, y^{i}\right)$ means there is a relation $y^{i}$ between the entity pair $\left(h_{k}^{i}, t_{k}^{i}\right)$ in instance $x_{k}^{i}$, and $r^{i}$ is corresponding relation description. $N$ is the number of relations and each relation with quite small $K$ labeled instances. For a FSIRC task, relation set $\mathcal{R}$ includes some pairs of inverse relations. For example`\emph{participant}' and ``\emph{participant of}" are inverse relations, and their relation descriptions are ``\emph{person or organization that actively takes part in the event}" and ``\emph{event a person or an organization was a participant in}", respectively.

\subsection{Overall Framework}
 As shown in Figure~\ref{img2}, our model consists of three parts:
\begin{itemize}
    \item \textbf{Instance Encoder} Given an instance and entity pair, we employ instance-level global encoder and phase-level local encoder to encode instance into an embedding. 
    \item \textbf{Function-words Enhanced Attention} Phase-level local encoder utilize function-words enhanced attention to capture important function words in instances.
    \item \textbf{Adaptive Message Passing} After computing embeddings, we transfer commonalities between same class instances and differences between different class instances.
\end{itemize}

\subsection{Instance Encoder}
Given an instance $x=\left\{w_{1}, \ldots, w_{l}\right\}$ mentioning two entities with $l$ words, we employ BERT \cite{devlin2019bert} as the encoder to obtain corresponding embeddings $\mathbf{X}=\left\{\mathbf{w}_{1}, \ldots, \mathbf{w}_{l}\right\}$, where each word embedding $\mathbf{w}_{i} \in \mathbb{R}^{d}$ and $d$ is embedding dimension. For $i$-th relation $r^{i}$, we feed the name and description into the BERT to obtain relation word embeddings $\mathbf{R}^{i} \in \mathbb{R}^{l \times d}$, and features of relations $\mathbf{r}^{i} \in \mathbb{R}^{2 d}$ are obtained by the hidden states corresponding to [CLS] token (converted to $2d$ dimension with a transformation).

For instance $x_{k}^{i}$ in $\mathcal{S}$ and query instance $x^{q}$, our model generates global instance embeddings and local phase embeddings to form hybrid instance embeddings $\mathbf{x}_{k}^{i}$ and $\mathbf{x}^{q}$. The following
takes $\mathbf{x}_{k}^{i}$ as an example to explain.

\paragraph{Instance-level Global Encoder} The global features $\left\{\overline{\mathbf{x}}_{k}^{i} \in \mathbb{R}^{2 d} ; i=1, \ldots, N, k=1, \ldots, K\right\}$ are obtained by concatenating the hidden states corresponding to start tokens of two entity mentions following \cite{soares2019matching}.

\paragraph{Phase-level Local Encoder} The main process consists of learning a keyword attention vector ${\boldsymbol{\alpha}}_{k}^{i} \in \mathbb{R}^{l}$ and a function-word attention vector ${\boldsymbol{\beta}}^{s} \in \mathbb{R}^{l}$. ${\boldsymbol{\alpha}}_{k}^{i}$ can be computed as follows:
\begin{small}
\begin{align}
    \setlength\abovedisplayskip{0.1cm} 
    \setlength\belowdisplayskip{0.1cm} 
    {\boldsymbol{\alpha}}_{k}^{i}=\operatorname{softmax}\left(\mathbf{X}_{k}^{i} * \mathbf{u}_{w}+\operatorname{sum}\left(\frac{\mathbf{X}_{k}^{i}\left(\mathbf{R}^{i}\right)^{T}}{\sqrt{d}}\right) / d\right)
\end{align}
\end{small}
where the memory unit $\mathbf{u}_{w} \in \mathbb{R}^{d}$ is a trainable parameter. It can help us to select general keywords from instances. 

\subsection{Function-words Enhanced Attention}
\label{p1}
%Inspired by meta-learning \cite{sun2019hierarchical}, 
We utilize class-general attention to learn general function words importance distribution and leverage class-specific attention consisting of constituent attention and semantic-related attention to estimate class-specific importance.

\paragraph{Class-general attention} We downweigh importance of words related to $\mathbf{u}_{w}$ and upweigh words importance irrelated to $\mathbf{u}_{w}$ to get general function-words importance ${\boldsymbol{\beta}}_{\text {general}} \in \mathbb{R}^{l}$, where $\mathbf{E} \in \mathbb{R}^{l \times 1}$ is an all-one matrix:
\begin{small}
\begin{align}
     \setlength\abovedisplayskip{0.1cm} 
    \setlength\belowdisplayskip{0.1cm}
    {\boldsymbol{\beta}}_{\text {general}}=\operatorname{softmax}\left(\mathbf{E}-\mathbf{X}_{k}^{i} \mathbf{u}_{w}\right)
\end{align}
\end{small}
\paragraph{Class-specific attention}
Considering function words importance varying by classes, we learn a constituent prior
matrix $\mathbf{C} \in \mathbb{R}^{l \times l}$ and semantic-related matrix $\mathbf{S} \in \mathbb{R}^{l \times l}$ and using them to strengthen attention of function words adjacent to keywords.

The element $C_{i, j}$ in $\mathbf{C}$ meaning the probability that $w_{i}$ and $w_{j}$ of instance $x_{k}^{i}$ belong to same phase, is obtained as follows, where $[\cdot]_{n}$ is the $n$-th row of a matrix :
\begin{small}
\begin{align}
    \setlength\abovedisplayskip{0.1cm} 
    \setlength\belowdisplayskip{0.1cm}
    s_{n, n+1} &=\frac{\left[\mathbf{X}_{k}^{i}\right]_{n}\times\left[\mathbf{X}_{k}^{i}\right]_{n+1}^{T}}{\sqrt{d}} \\
    p_{n, n+1}, p_{n, n-1} &=\operatorname{softmax}\left(s_{n, n+1}, s_{n, n-1}\right) \\
    a_{n} =\sqrt{p_{n, n+1} \times p_{n+1, n}} \quad & \quad
    C_{i, j} =e^{\sum_{n=i}^{j-1} \log \left(a_{n}\right)} 
\end{align}
\end{small}

We compute the score $s_{n, n+1}$ representing the tendency that $w_{n}$ and right neighbor $w_{n+1}$ belong to the same phase by scaled dot-product attention. We constrain it to either belong to its right neighbor or left neighbor, which is implemented by applying a softmax function to two attention links of $w_{n}$. As $p_{n, n+1}$ and $p_{n+1, n}$ possibly have different values, we average its two attention links.

Then we use a self-attention mechanism to obtain semantic-related matrix $\mathbf{S}$ to attend necessary function words far from keywords:
\begin{small}
\begin{align}
\setlength\abovedisplayskip{0.1cm} 
\setlength\belowdisplayskip{0.1cm}
  \mathbf{S}=\frac{\mathbf{X}_{k}^{i}\times\left(\mathbf{X}_{k}^{i}\right)^{T}}{\sqrt{d}}
\end{align}
\end{small}
Next we find the keywords index $\mathbf{I}_{k}^{i}$ in $x_{k}^{i}$ and strengthen related function words from matrix $\mathbf{C}$ and $\mathbf{S}$ according to index, $\max (\cdot)_{r}$ is used to get indexes of the top-$r$ largest attention keyword for a matrix where $r$ is the number of keywords:
\begin{small}
\begin{align}
    \setlength\abovedisplayskip{0.1cm} 
    \setlength\belowdisplayskip{0.1cm}
    \mathbf{I}_{k}^{i}=\max \left({\boldsymbol{\alpha}}_{k}^{i}\right)_{r}\quad
    {\boldsymbol{\beta}}_{\text {constituent }}=[\mathbf{C}]_{\mathbf{I}_{k}^{i}} \quad
    {\boldsymbol{\beta}}_{\text {related }}=[\mathbf{S}]_{\mathbf{I}_{k}^{i}}
\end{align}
\end{small}

Finally, the model uses ${\boldsymbol{\beta}}_{\text {constituent }}$,${\boldsymbol{\beta}}_{\text {related}}$ and ${\boldsymbol{\beta}}_{\text {general}}$ to form hybrid function-word attention vector ${\boldsymbol{\beta}}$, where $\lambda$ is hyper-parameter:
\begin{small}
\begin{align}
    \setlength\abovedisplayskip{0.1cm} 
    \setlength\belowdisplayskip{0.1cm}
    {\boldsymbol{\beta}}^{\prime} &=\lambda{\boldsymbol{\beta}}_{\text {constituent}}+(1-\lambda) {\boldsymbol{\beta}}_{\text {related}} \\
    {\boldsymbol{\beta}} &=\frac{1}{r} \sum_{i=1}^{r}\left[{\boldsymbol{\beta}}^{\prime}\right]_{i}+{\boldsymbol{\beta}}_{\text {general}} 
\end{align}
\end{small}
All in all, inspired by MAML\cite{finn2017model} learning general model parameters and fine-tunning them to adapt specific class, we design class-general and class-specific attention to learn function-words variance in few-shot setting.

\subsection{Adaptive Message Passing}
Message Passing model is used to reduce intra-class redundancy and adaptively control proportion of transferred inter-class message. Firstly, we construct a directed graph $G=(\mathbf{V}, \mathbf{E})$, where $\mathbf{V}=\left[\mathbf{x}_{1}^{1}; \ldots ; \mathbf{x}_{N}^{K}\right]$ is a set of instances features with $|V|=N \times K$ and $E$ is adjacency matrix, where $[;]$ denotes row-wise concatenation, $\mathbf{v}^{i} \in \mathbb{R}^{d}$ denotes the $i$-th row of $\mathbf{V}$.
\begin{small}
\begin{align}
    \setlength\abovedisplayskip{0.05cm} 
    \setlength\belowdisplayskip{0.05cm}
    E_{i j}=\left\{\begin{array}{cc}
    \frac{\left(\mathbf{v}^{i}\right)^{T} \mathbf{v}^{j}}{\left\|\mathbf{v}^{i}\right\|_{2}\left\|\mathbf{v}^{j}\right\|_{2}} & \text { if } i \neq j \\
    0 & \text { if } i=j
    \end{array}\right.
\end{align}
\end{small}

 We design a new node updating way that captures and transfers inter-class differences and intra-class commonalities between instance nodes: where $\mathcal{N}_{i}^{0}$ and $\mathcal{N}_{i}^{1}$ denote the different and same class neighbor set as $\mathbf{v}^{i}$, respectively. $d_{i}$ is the degree of instance node $i$.
\begin{small}
\begin{equation}
\begin{array}{l}
    \setlength\abovedisplayskip{0.05cm} 
    \setlength\belowdisplayskip{0.05cm}
    \tilde{\mathbf{v}}^{i}=\mathbf{v}^{i}-\sum_{j \in \mathcal{N}_{i}^{0}} e_{i} \mathbf{v}^{j}+\sum_{j \in \mathcal{N}_{i}^{1}} e_{i} \mathbf{v}^{j} \\
    e_{i}=\frac{\max \left(e_{i}^{j}-\bar{e}_{i}, 0\right)}{\sqrt{d_{i} d_{j}}} \\
    \bar{e}_{i}=\frac{1}{2\left(N^{*} K-1\right)}\left(\sum_{j \in \mathcal{N}_{i}^{0}} e_{i}^{j}+\sum_{j \in \mathcal{N}_{i}^{1}} e_{i}^{j}\right)
\end{array}
\end{equation}
\end{small}

For $i$-th relation, we average $K$ supporting features to form prototype representation $\mathbf{p}^{i}$ following \cite{snell2017prototypical}.
\begin{small}
\begin{align}
    \setlength\abovedisplayskip{0.1cm} 
    \setlength\belowdisplayskip{0.1cm}
    \mathbf{p}^{i}=\frac{1}{K} \sum_{j=(i-1) \times K}^{i\times K} \mathbf{v}^{j}
\end{align}
\end{small}
With $N$ prototype relations, the model computes the probability of the relations for the query instance $x^{q}$ as follows:
\begin{small}
\begin{align}
    \setlength\abovedisplayskip{0.1cm} 
    \setlength\belowdisplayskip{0.1cm}
    z\left(y=i \mid x_{q}\right)=\frac{\exp \left(\mathbf{x}^{q} \cdot \mathbf{p}^{i}\right)}{\sum_{n=1}^{N} \exp \left(\mathbf{x}^{q} \cdot \mathbf{p}^{n}\right)}
\end{align}
\end{small}

The final objective function is formally written as: $\mathcal{L}_{C E}=-\log \left(z_{y}\right)$, where $y$ is  class label, and $z_{y}$ is estimated probability for the class $\mathrm{y}$.
%\begin{small}
%\begin{align}
%    \mathcal{L}_{C E}=-\log \left(z_{y}\right)
%\end{align}
%\end{small}

In a short, we design a new node updating method to capture inter-class differences additionally.

\subsection{Theoretical Analysis}
In our theoretical analysis, we show that the involvement of function words will increase intra-class differences(Theorem\ref{t1}) and the designed message passing mechanism makes different class nodes become discriminative and same class nodes similar.(Theorem\ref{t2}) Detailed theoretical proof is shown in the Appendix.

Given any two instances $x_{i}$ and $x_{j}$, let the corresponding keywords representations be $\mathbf{x}_{i}^{c}=\left\{a_{i 1}, a_{i 2}, \ldots, a_{i d}\right\} \in \mathbb{R}^{d}$  and 
$\mathbf{x}_{j}^{c}=\left\{b_{j 1}, b_{j 2}, \ldots, b_{j d}\right\} \in \mathbb{R}^{d}$;  the function words representations be $\mathbf{x}_{i}^{f}=\left\{a_{i}^{1}, a_{i}^{2}, \ldots, a_{i}^{d}\right\} \in \mathbb{R}^{d}$ and
$\mathbf{x}_{j}^{f}=\left\{b_{j}^{1}, b_{j}^{2}, \ldots, b_{j}^{d}\right\} \in \mathrm{R}^{d}$; 
and the instance representations considering function words be $\mathbf{x}_{i}=\mathbf{x}_{i}^{c}+\mathbf{x}_{i}^{f}$ and  $\mathbf{x}_{j}=\mathbf{x}_{j}^{c}+\mathbf{x}_{j}^{f}$.\\

\begin{theorem}
Let \begin{small}
\begin{align}
    \setlength\abovedisplayskip{0.05cm}
    \setlength\belowdisplayskip{0.05cm}
    \operatorname{norm}(x_{i} \odot x_{j})=\frac{\left|x_{i} \odot x_{j}\right|}{\left\|x_{i} \odot x_{j}\right\|_{2}} 
\end{align}
\end{small}
where $\odot$ indicates the inner product of vectors.

If $\operatorname{norm}(x_{i}^{f} \odot x_{j}^{f}) \leq \operatorname{norm}(x_{i}^{c} \odot x_{j}^{c})$, then
\begin{small}
\begin{align}
    \setlength\abovedisplayskip{0.1cm}
    \setlength\belowdisplayskip{0.1cm}
    \operatorname{norm}(x_{i} \odot x_{j}) \leq \operatorname{norm}(x_{i}^{c} \odot x_{j}^{c})
\end{align}
\end{small}
\label{t1}
\end{theorem}

This theorem shows that when function words are taken into account, the similarity degree ${norm}(x_{i} \odot x_{j})$ between instances $x_{i}$ and $x_{j}$ becomes smaller and thus the intra-class difference is increased.

\begin{theorem}
Given any two function words $\mathbf{x}_{i}$ and $\mathbf{x}_{j}$ and their instance representations, define the similarity measure between $\mathbf{x}_{i}$ and $\mathbf{x}_{j}$ as
\begin{small}
\begin{align}
    \setlength\abovedisplayskip{0.1cm}
    \setlength\belowdisplayskip{0.1cm}
D\left(\mathbf{x}_{i}, \mathbf{x}_{j}\right)=\mathbf{x}_{i} \odot \mathbf{x}_{j} 
\end{align}
\end{small}
The message passings between same class instances and different classes are defined as, respectively
\begin{small}
\begin{align}
    \setlength\abovedisplayskip{0.1cm}
    \setlength\belowdisplayskip{0.1cm}
\mathbf{x}_{i}^{\prime}=\mathbf{x}_{i}+\operatorname{norm}\left(\mathbf{x}_{i} \odot \mathbf{x}_{j}\right) \mathbf{x}_{j}\\
\mathbf{x}_{i}^{\prime}=\mathbf{x}_{i}-\operatorname{norm}\left(\mathbf{x}_{i} \odot \mathbf{x}_{j}\right) \mathbf{x}_{j}
\end{align}
\end{small}
If $\mathbf{x}_{i}$ and $\mathbf{x}_{i}$ belong to the same class, then
\begin{small}
\begin{align}
    \setlength\abovedisplayskip{0.1cm}
    \setlength\belowdisplayskip{0.1cm}
D\left(\mathbf{x}_{i}, \mathbf{x}_{j}\right) \leq D\left(x_{i}^{\prime}, \mathbf{x}_{j}^{\prime}\right)
\end{align}
\end{small}
If $\mathbf{x}_{i}$ and $\mathbf{x}_{i}$ belong to different classes, then
\begin{small}
\begin{align}
    \setlength\abovedisplayskip{0.1cm}
    \setlength\belowdisplayskip{0.1cm}
D\left(\mathbf{x}_{i}, \mathbf{x}_{j}\right) \geq D\left(\mathbf{x}_{i}^{\prime}, \mathbf{x}_{j}^{\prime}\right)
\end{align}
\end{small}
\label{t2}
\end{theorem}
This result shows that, compared with the original similarity $D\left(\mathbf{x}_{i}, \mathbf{x}_{j}\right)$, if $D\left(\mathbf{x}_{i}^{\prime}, \mathbf{x}_{j}^{\prime}\right)$  becomes smaller, then the message passing transfer transfers different information, otherwise it transfers similar information.

\section{Experiments}
\begin{table*}[t]
\centering
\setlength{\belowcaptionskip}{-4mm}
\renewcommand{\arraystretch}{0.90}
\setlength{\tabcolsep}{5mm}{
\begin{small}
\begin{tabular}{clcccc}
\hline
\small{Encoder}  & \small{Model} & \small{5-way-1-shot} & \small{5-way-5-shot} & \small{10-way-1-shot} & \small{10-way-5-shot} \\
\hline
\multirow{3}*{CNN}&
Proto-CNN \cite{snell2017prototypical} $\triangleright$ & 72.65 / 74.52& 	86.15 / 88.40	& 60.13 / 62.38	& 76.20 / 80.45\\&
Proto-HATT \cite{gao2019hybrid}&	75.01 / -- --&	87.09 / 90.12&	62.48 / -- --	&77.50 / 83.05\\
&MLMAN \cite{ye2019multi} $\diamond$ &	78.85 / 82.98&88.32 / 92.66&	67.54 / 73.59&79.44 / 87.29\\
\hline
\multirow{13}*{Bert}&
Proto-Bert \cite{snell2017prototypical} $\circ$ &\small{	82.92 / 80.68}&	\small{91.32 / 89.60}	&\small{73.24 / 71.48}&	\small{83.68 / 82.89}\\&
\small{MAML \cite{finn2017model} $\circ$} &	\small{82.93 / 89.70}	&\small{ 86.21 / 93.55}&	\small{73.20 / 83.17}&	\small{76.06 / 88.51}\\&
\small{GNN} \cite{garcia2017few}&	\small{74.21 / 75.66}	&\small{86.16 / 89.06}&	\small{67.98 / 70.08}&	\small{73.65 / 76.93}\\&
\small{BERT-PAIR \cite{gao2019fewrel} $\triangleright$ }&	\small{85.66 / 88.32}&	\small{89.48 / 93.22}	&\small{76.84 / 80.63}&	\small{81.76 / 87.02}\\&
\small{REGRAB \cite{qu2020few} $\circ$ }&	\small{87.93 / 90.30}&	\small{92.58 / 94.25}&\small{	80.52 / 84.09	}&\small{87.02 / 89.93}\\&
\small{TD-Proto} \cite{sun2019hierarchical}&	\small{83.43 / 84.53}&	\small{90.26 / 92.38}&\small{	72.45 / 74.32}&	\small{82.10 / 85.19}\\&
\small{ConceptFERE} \cite{yang2021entity}&	\small{87.21 / 89.21}	&\small{90.53 / 93.98}&\small{	73.56 / 75.72}&	\small{83.29 / 86.21}\\&
TPN \cite{wen2021enhanced}&	\small{-- -- / 80.14}&	\small{-- -- / 93.60}	&\small{-- -- / 72.67}	&\small{-- -- / 89.83}\\&
\small{CTEG \cite{wang2020learning} $\diamond$} &	\small{84.72 / 88.11}&\small{	92.52 / 95.25}&	\small{76.01 / 81.29}	&\small{84.89 / 91.33}\\
&
\small{\textbf{FAEA(ours)+AMP}}&	\small{\textbf{90.81 / 95.10}}	&\small{\textbf{94.24 / 96.48}}&	\small{\textbf{84.22 / 90.12}}&\small{	\textbf{88.74 / 92.72}}\\
\hline &
\small{MTB \cite{soares2019matching} $\triangleright$} &	\small{-- -- / 93.86}&	\small{-- -- / 97.06}&	\small{-- --/89.20}	& \small{--  -- / 94.27}\\&
\small{CP} \cite{peng2020learning}&\small{	-- -- / 95.10}&	\small{-- -- / 97.10}&	\small{-- -- / 91.20}&	\small{-- -- / 94.70}\\&
\small{\textbf{FAEA+CP}}&	\small{\textbf{94.11 / 96.36}}	&\small{\textbf{89.55 / 97.85}}&	\small{\textbf{86.59 / 93.82}}&\small{	\textbf{93.64 / 96.29}}\\
\hline
\end{tabular}
\end{small}
}
\caption{Accuracy (\%) of few-shot classification on the FewRel 1.0 validation / test set.}
\label{tab1}
\end{table*}
\begin{table}[tb]
\centering
\setlength{\belowcaptionskip}{-5mm}
\renewcommand{\arraystretch}{0.93}
\begin{small}
\begin{tabular}{lcccc}
\hline
	Model &\tabincell{c}{5-way\\1-shot}
	&\tabincell{c}{5-way\\5-shot}
	&\tabincell{c}{10-way\\1-shot}
	&\tabincell{c}{10-way\\5-shot} \\
\hline
Proto-CNN&35.09&	49.37&	22.98&35.22\\
\small{Proto-BERT}&	40.12&51.50&	26.45&36.93\\
\small{Proto-ADV}&42.21&58.71& 28.91&44.35 \\
\small{Bert-Pair}&67.41&\small{78.57}	&\small{54.89}&66.85\\
\textbf{Our}&\textbf{73.58}&\textbf{90.10}&\textbf{62.98}&\textbf{80.51}\\
\hline
\end{tabular}
\end{small}
\caption{Accuracy (\%) of few shot classification on the FewRel2.0 domain adaptation test set.}
\label{tab2}
\end{table}
\begin{table}[tb]
\setlength{\belowcaptionskip}{-4mm}
\renewcommand{\arraystretch}{0.85}
\setlength{\tabcolsep}{0.2mm}{

\begin{tabular}{lcccccccccc} 
\multicolumn{6}{c}{}\\  %横向合并7列单元格  两侧添加竖线
\hline
\multirow{2}*{\footnotesize Model}&
\multicolumn{1}{c}{\footnotesize 2way1shot}&
\multicolumn{1}{c}{\footnotesize 4way1shot}
&\multicolumn{1}{c}{\footnotesize 5way1shot}
&\multicolumn{1}{c}{\footnotesize 5way3shot}
&\multicolumn{1}{c}{\footnotesize 5way5shot}
\\  &\footnotesize R - I&\footnotesize R - I&\footnotesize R - I&\footnotesize R - I&\footnotesize R - I\\
\hline
\multirow{1}*{\footnotesize Proto-HATT}
&\scriptsize83.26 53.62&\scriptsize78.61 49.72&\scriptsize75.01 62.13
&\scriptsize80.53 68.15&\scriptsize87.09 73.02
\\
\multirow{1}*{\footnotesize Proto-Bert}
&\scriptsize 87.54 54.96 &\scriptsize83.44 51.25&\scriptsize82.92 66.23&\scriptsize
85.23 67.96&\scriptsize91.32 72.53
\\
\multirow{1}*{\footnotesize Bert-Pair}
&\scriptsize91.21 56.20&\scriptsize87.44 54.87&\scriptsize85.66 67.53&\scriptsize88.42 69.92&\scriptsize89.48 71.21\\
\multirow{1}*{\footnotesize TD-Proto}
&\scriptsize89.69 53.81&\scriptsize85.36 52.31&\scriptsize83.25 63.21
&\scriptsize84.21 65.32 &\scriptsize85.21 70.19\\

\multirow{1}*{\footnotesize ConceptFERE}
&\scriptsize92.57	62.21&\scriptsize88.89	59.76&\scriptsize87.21	69.47
&\scriptsize88.79	71.15&\scriptsize	90.53	76.24\\
		
\multirow{1}*{\footnotesize \textbf{Our}}
&\scriptsize\textbf{97.65	78.96}&\scriptsize\textbf{92.21	75.45}&\scriptsize\textbf{90.81	80.02}
&\scriptsize\textbf{91.96	82.26} &\scriptsize\textbf{94.24	85.63}\\
\hline
\end{tabular}}
\caption{Accuracy (\%) of different few-shot settings on FewRel1.0. `R' stands for the standard few-shot setting and `I' stands for evaluating including inverse relations.}
\label{tab3}
\end{table}
\begin{table}[t]
\centering
\setlength{\belowcaptionskip}{-4mm}
\renewcommand{\arraystretch}{0.85}
\begin{small}
\begin{tabular}{lccc}
\hline
	Model &Id &\tabincell{c}{5-way\\1-shot} &\tabincell{c}{10-way\\1-shot} \\
\hline
\small{\textbf{Our}}&1	&	\textbf{90.81}&	\textbf{84.22}\\
\small{-- phase-level encoding}&2&	84.98&	77.02\\
\small{-- function word attn}&	\small{3}&\small{87.52}&	\small{80.92}\\
\small{-- general attn}&\small{4	}&\small{88.81}& \small{82.08} \\
\small{-- constituent attn}&\small{5	}&\small{88.93}	&\small{82.69}\\
\small{-- related attn}&6&89.43&	83.51\\
\small{-- message passing}&7	&90.01&	83.62	\\
\small{-- mean }&8&	90.32&	83.86	\\
\hline
\end{tabular}
\end{small}
\caption{Ablation study on FewRel 1.0 validation set showing accuracy (\%).}
\label{tab4}
\end{table}
\subsection{Baselines} 
\label{p5}
\begin{itemize}
    \item To demonstrate the usefulness of local-level features, compare with
    global-level models:
    metric-based models
    \textbf{Proto} \cite{snell2017prototypical}, \textbf{Proto-HATT} \cite{gao2019hybrid}, \textbf{MLMAN} \cite{ye2019multi}, \textbf{BERT-PAIR} \cite{gao2019fewrel} and \textbf{TPN} \cite{wen2021enhanced}. And gradient-based models,  \textbf{MAML} \cite{finn2017model} and \textbf{GNN} \cite{garcia2017few}.
    \item To prove function words importance, compare with word-level models: \textbf{TD-Proto} \cite{sun2019hierarchical}, usinging memory network to learn general content word importance distribution. \textbf{ConceptFERE} \cite{yang2021entity},designing an attentive model to measure the importance of each word for a specific class.
    \item Models introducing external information. \textbf{REGRAB} \cite{qu2020few} utilizes  relation graph knowledge. \textbf{CTEG} \cite{wang2020learning} uses dependency trees. 
    \item Pretrained RE methods: \textbf{MTB} \cite{soares2019matching}, pretrain with their proposed matching the blank task and \textbf{CP} \cite{peng2020learning}, an entity masked contrastive pretraining framework. 
\end{itemize}

\subsection{Datasets and Settings}  
We evaluate our model on \textbf{FewRel1.0} \cite{han2018fewrel} and \textbf{FewRel2.0} \cite{gao2019fewrel}, consisting of 100 relations and each with 700 labeled instances. Our experiments follow the splits used in official benchmarks, which split the dataset into 64 base classes for training, 16 classes for validation, and 20 classes for testing.

We evaluate our model in terms of the averaged accuracy on query set of multiple N-way-K-shot tasks. According to  \cite{gao2019hybrid,gao2019fewrel}, we choose $N$ to be 5 and 10, and $K$ to be 1 and 5 to form 4 scenarios. In addition, we take base-uncased BERT as the encoder of 768 dimensions for fair comparison. The input max length is set to 128. Besides, the AdamW optimizer is applied with the learning rate and weight decay as $2 \times 10^{-5}$ and $1 \times 10^{-2}$. Furthermore, hyper-parameter $\lambda$ is set to $0.6$ and $\mathbf{u}_{w}$ is randomly initialized following \cite{sun2019hierarchical}. 

\subsection{Results}
\setlength{\belowcaptionskip}{-4mm}
\begin{figure}[t]
\centering
\includegraphics[height=5cm]{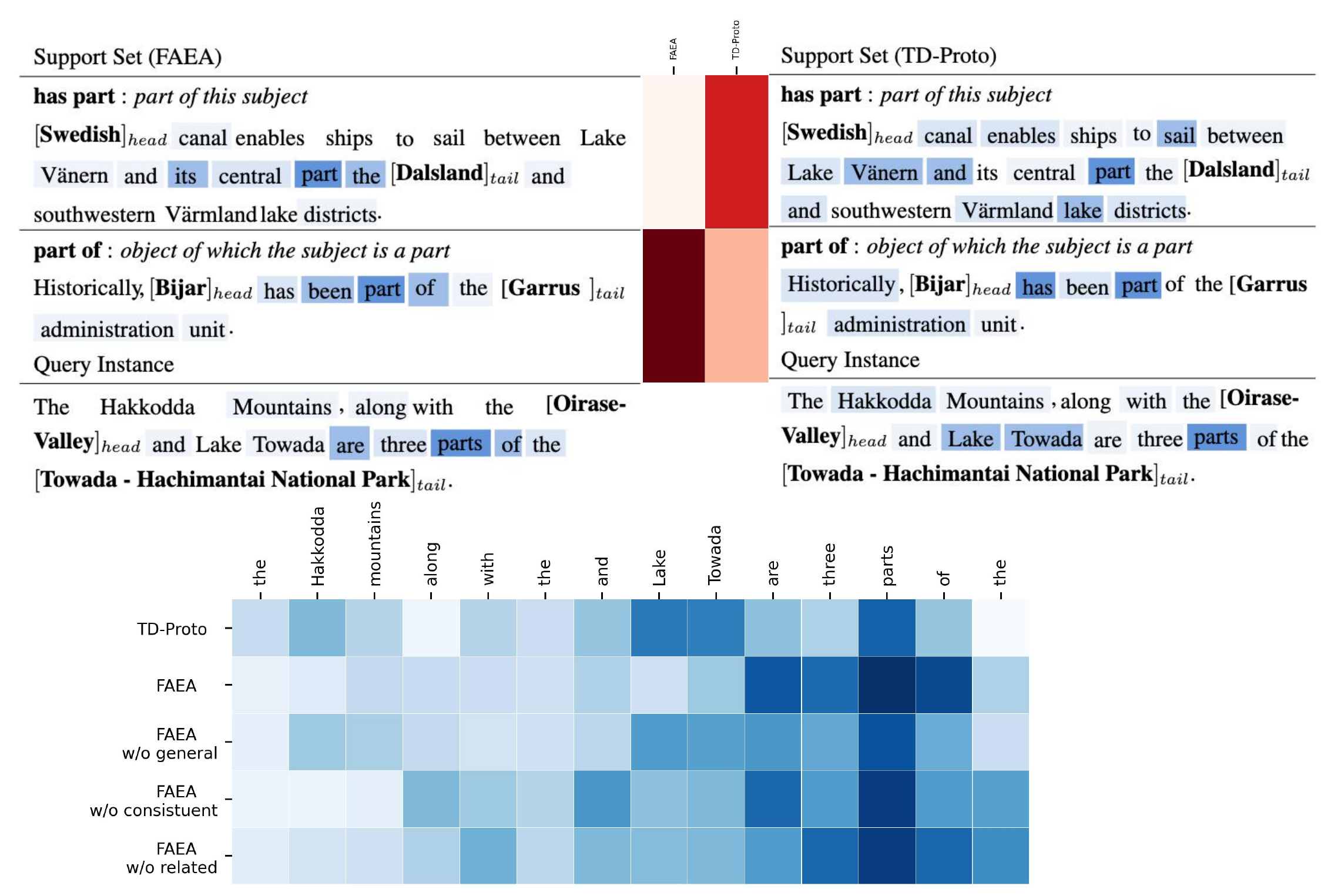}
\caption{A 2-way-1-shot inverse relation task example. The upper part visualizes the attention score of each word by FAEA and TD-Proto, the middle part is similarities between the query instance and support instances. The lower part is attention scores of words in query under different models.}
\label{fig2}
\end{figure} 
\begin{figure}[ht]
\centering
\setlength{\belowcaptionskip}{-6mm}
\includegraphics[scale=.28]{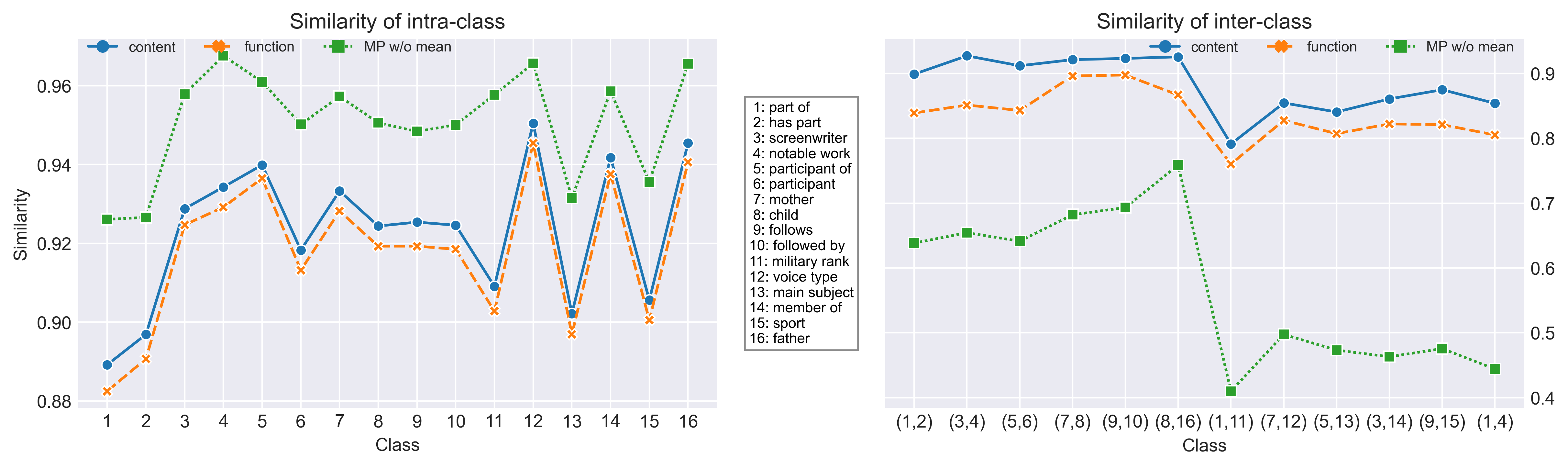}
\caption{The similarity of intra-class and inter-class between some classes computed by dot-product }
\label{fig3}
\end{figure}
\begin{figure}[h]
\centering
\setlength{\belowcaptionskip}{-5mm}
\includegraphics[scale=.4]{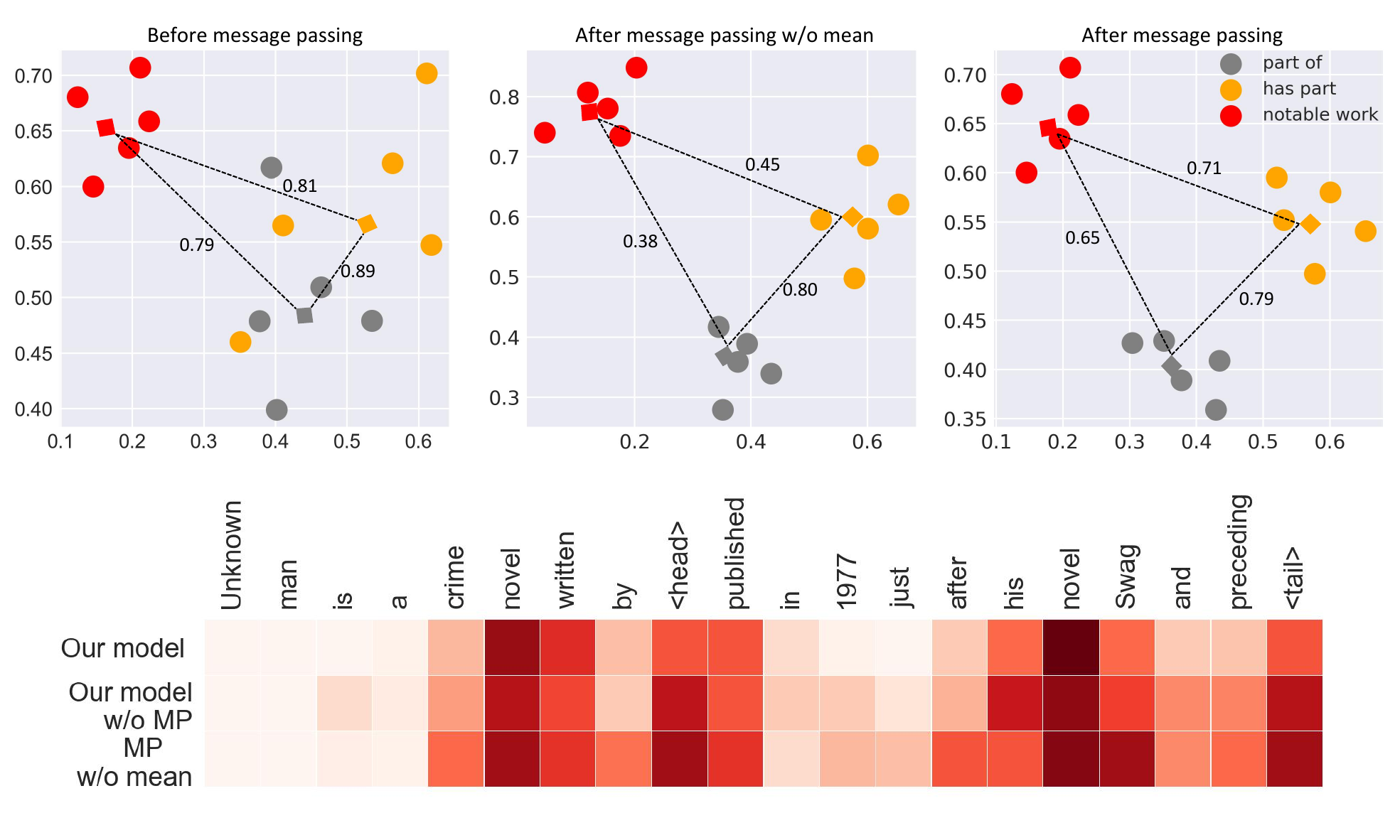}
\caption{The upper part is T-SNE plots of instance embeddings of ''\emph{notable work}'',''\emph{has part}'' and ''\emph{part of}'' with or without mean. The lower part is attention scores of words in ''\emph{notable work}'' under different models.}
\label{fig4}
\end{figure} 

\paragraph{Performance on FSRC}
 As shown in the upper part of Table \ref{tab1}, our method outperforms the strong baseline models by a large margin, especially in 1-shot scenarios. Specifically, we improve 5-way-1-shot and 10-way-1-shot tasks $4.80$ points and $6.06$ points in terms of accuracy, demonstrating the superior ability. In addition, our method achieves good performance on FewRel2.0, as shown in Table \ref{tab2}.
\begin{itemize}
    \item Proto and GNN, as widely-used baselines for few-shot learning, perform not well on FSRC. Unlike low-level patterns can be shared across tasks in computer vision, words that are informative for one task may not be relevant for other tasks. But these models ignore such local words importance variations learning. But FAEA leverages phase-level attention to attend local features.\item TD-PROTO and ConceptFERE also use semantic-level attention to explore content words, but neglect function words maintaining syntactic structure differences. Since FAEA captures function words to form fine-grained features, it obtains better performance. \item When computing relation prototypes, Proto-HATT and TPN utilize intra-class commonalities not considering inter-class differences. FAEA captures and leverages differences to get more discriminative representations. 
\end{itemize}
\paragraph{Performance on FSIRC}
To further illustrate the model effectiveness for FSIRC, we evaluate models on FewRel1.0 validation set with different settings, as shown in Table \ref{tab3}. Random is general evaluation setting, which samples 10,000 test tasks randomly from validation relations. Inverse represents each evaluated task includes inverse relations. As we can see, the baselines achieve good accuracy under random settings but  drops significantly under inverse settings, around 26.98 points in 1-shot scenarios, which illustrates that FSIRC tasks are extremely challenging. FAEA gains the best accuracy, especially under inverse settings, proving that it can effectively capture function words and handle FSIRC tasks.

\section{Analysis}
\subsection{Analysis of function words attention}
This section discusses the effect of function-words attention. As shown in the Table \ref{tab4}, removing phase-level (Model 2) and function words attention (Model 3) severely decrease the performance, indicating function words are also essential to represent relations. Furthermore, as shown in Figure \ref{fig2}, with the help of function words attention, we highlight `\emph{are}' and `\emph{of}' to form the phase ``\emph{are part of}'', which appears in query and support instance of class ``\emph{part of}'' at the same time, then this support instance get higher similarity, and our model correctly classifies the query.

To demonstrate the effectiveness of three components of function-words attention, from model 4,5,6 of Table \ref{tab4}, we can see there is a performance decline if removing three components separately. As shown in lower part of Figure \ref{fig2}, we find that TD-Proto mainly attends content words such as `\emph{parts}',`\emph{Lake}'. FAEA without general attention not only enhances function words importance but also content words irrelated to keywords, such as `\emph{mountains}',`\emph{Hakkodda}'. And FAEA without constituent attention enhances some keywords-irrelated function words such as `\emph{along}', `\emph{and}'. FAEA without related attention tends to decrease some related function words importance far away from the keywords such as `\emph{are}'. FAEA further captures correct function words to form ``\emph{are three parts of}'' , which demonstrates that three components all contribute to enhance function words importance.
 
\subsection{Analysis of message passing}As shown in Table \ref{tab4}, we compare models without message passing mechanism (Model 7) and message passing without mean (Model 8). We observe that considering message passing achieves higher accuracy, and adding the mean to control the proportion of transferred inter-class message further improves the performance.

To futher demonstrate the effectiveness of message passing, we choose some classes and visualize the similarity shown in Figure \ref{fig3}. We can see that only considering content words, the inter-class information of inverse relations has a high similar score. And the introduction of function words can effectively reduce it. But we also find function words reduce the similarity of intra-class information from left part. The designed message passing mechanism without mean can effectively increase intra-class commonalities and keep inter-class differences. But from right part, we observe when the inter-class differences are large enough, with the message passing mechanism, the similarity of the inter-class sharp decline, it will destroy original relation semantic. 

Specifically, from upper part of Figure \ref{fig4}, we can see instances of ``\emph{part of}'' and ``\emph{notable work}'' are quite different with similarity of 0.79, much lower than the score 0.89 between ``\emph{has part}'' and ``\emph{part of}''. But after message passing without mean, their similarity sharply decrease to 0.38 and we can see the instance of ``\emph{notable work}'' attends some distinct but class-irrelated words from lower part of figure \ref{fig4}, such as `\emph{crime}',`\emph{by}'. After message passing with mean, the similarity is slowly decline to 0.65 and we can see it enhances class-related words ‘novel’ and avoids introducing noise. All results show that the best performance is achieved by message passing with mean.

\section{Conclusion}
In this paper, we have presented FAEA, a framework that can effectively handle few-shot inverse relations by enhancing related function words importance. Experiments demonstrate that FAEA achieves new sota results on two NLP tasks on FewRel dataset. In future work, we will try to design a more effective and general function-words enhanced backbone network for various NLP tasks.

\newpage
\bibliography{ijcai22}

\begin{thebibliography}{}

\bibitem[\protect\citeauthoryear{Bao \bgroup \em et al.\egroup
  }{2020}]{bao2019few}
Yujia Bao, Menghua Wu, Shiyu Chang, and Regina Barzilay.
\newblock Few-shot text classification with distributional signatures.
\newblock In {\em ICLR}, Addis Ababa, Ethiopia, April 2020. OpenReview.net.

\bibitem[\protect\citeauthoryear{Devlin \bgroup \em et al.\egroup
  }{2019}]{devlin2019bert}
Jacob Devlin, Ming{-}Wei Chang, Kenton Lee, and Kristina Toutanova.
\newblock {BERT:} pre-training of deep bidirectional transformers for language
  understanding.
\newblock In {\em NAACL-HLT}, pages 4171--4186, Minneapolis, MN, USA, June
  2019. Association for Computational Linguistics.

\bibitem[\protect\citeauthoryear{Finn \bgroup \em et al.\egroup
  }{2017}]{finn2017model}
Chelsea Finn, Pieter Abbeel, and Sergey Levine.
\newblock Model-agnostic meta-learning for fast adaptation of deep networks.
\newblock In {\em ICML}, pages 1126--1135, Sydney, NSW, Australia, August 2017.
  {PMLR}.

\bibitem[\protect\citeauthoryear{Gao \bgroup \em et al.\egroup
  }{2019a}]{gao2019hybrid}
Tianyu Gao, Xu~Han, Zhiyuan Liu, and Maosong Sun.
\newblock Hybrid attention-based prototypical networks for noisy few-shot
  relation classification.
\newblock In {\em AAAI}, pages 6407--6414, Honolulu, Hawaii, USA,
  January--February 2019. {AAAI} Press.

\bibitem[\protect\citeauthoryear{Gao \bgroup \em et al.\egroup
  }{2019b}]{gao2019fewrel}
Tianyu Gao, Xu~Han, Hao Zhu, Zhiyuan Liu, Peng Li, Maosong Sun, and Jie Zhou.
\newblock Fewrel 2.0: Towards more challenging few-shot relation
  classification.
\newblock In {\em EMNLP-IJCNLP}, pages 6249--6254, Hong Kong, China, November
  2019. Association for Computational Linguistics.

\bibitem[\protect\citeauthoryear{Han \bgroup \em et al.\egroup
  }{2018}]{han2018fewrel}
Xu~Han, Hao Zhu, Pengfei Yu, Ziyun Wang, Yuan Yao, Zhiyuan Liu, and Maosong
  Sun.
\newblock Fewrel: {A} large-scale supervised few-shot relation classification
  dataset with state-of-the-art evaluation.
\newblock In {\em EMNLP}, pages 4803--4809, Brussels, Belgium,
  October--November 2018. Association for Computational Linguistics.

\bibitem[\protect\citeauthoryear{Ma \bgroup \em et al.\egroup
  }{2021}]{Kaixin2021leveragin}
Kaixin Ma, Filip Ilievski, Jonathan Francis, Yonatan Bisk, Eric Nyberg, and
  Alessandro Oltramari.
\newblock Knowledge-driven data construction for zero-shot evaluation in
  commonsense question answering.
\newblock In {\em AAAI}, pages 13507--13515, Online Event, February 2021.
  {AAAI} Press.

\bibitem[\protect\citeauthoryear{Peng \bgroup \em et al.\egroup
  }{2020}]{peng2020learning}
Hao Peng, Tianyu Gao, Xu~Han, Yankai Lin, Peng Li, Zhiyuan Liu, Maosong Sun,
  and Jie Zhou.
\newblock Learning from context or names? an empirical study on neural relation
  extraction.
\newblock In {\em EMNLP}, pages 3661--3672, Online, November 2020. Association
  for Computational Linguistics.

\bibitem[\protect\citeauthoryear{Qu \bgroup \em et al.\egroup
  }{2020}]{qu2020few}
Meng Qu, Tianyu Gao, Louis{-}Pascal A.~C. Xhonneux, and Jian Tang.
\newblock Few-shot relation extraction via bayesian meta-learning on relation
  graphs.
\newblock In {\em ICML}, pages 7867--7876, Virtual Event, July 2020. {PMLR}.

\bibitem[\protect\citeauthoryear{Satorras and Estrach}{2018}]{garcia2017few}
Victor~Garcia Satorras and Joan~Bruna Estrach.
\newblock Few-shot learning with graph neural networks.
\newblock In {\em ICLR}, Vancouver, BC, Canada, April--May 2018.
  OpenReview.net.

\bibitem[\protect\citeauthoryear{Snell \bgroup \em et al.\egroup
  }{2017}]{snell2017prototypical}
Jake Snell, Kevin Swersky, and Richard~S. Zemel.
\newblock Prototypical networks for few-shot learning.
\newblock In {\em NIPS}, pages 4077--4087, Long Beach, CA, {USA}, December
  2017. Advances in Neural Information Processing Systems.

\bibitem[\protect\citeauthoryear{Soares \bgroup \em et al.\egroup
  }{2019}]{soares2019matching}
Livio~Baldini Soares, Nicholas FitzGerald, Jeffrey Ling, and Tom Kwiatkowski.
\newblock Matching the blanks: Distributional similarity for relation learning.
\newblock In {\em ACL}, pages 2895--2905, Florence, Italy, July--August 2019.
  Association for Computational Linguistics.

\bibitem[\protect\citeauthoryear{Sun \bgroup \em et al.\egroup
  }{2019}]{sun2019hierarchical}
Shengli Sun, Qingfeng Sun, Kevin Zhou, and Tengchao Lv.
\newblock Hierarchical attention prototypical networks for few-shot text
  classification.
\newblock In {\em EMNLP-IJCNLP}, pages 476--485, Hong Kong, China, November
  2019. Association for Computational Linguistics.

\bibitem[\protect\citeauthoryear{Wang \bgroup \em et al.\egroup
  }{2020}]{wang2020learning}
Yuxia Wang, Karin Verspoor, and Timothy Baldwin.
\newblock Learning from unlabelled data for clinical semantic textual
  similarity.
\newblock In {\em ClinicalNLP}, pages 227--233, Online, November 2020.
  Association for Computational Linguistics.

\bibitem[\protect\citeauthoryear{Wen \bgroup \em et al.\egroup
  }{2021}]{wen2021enhanced}
Wen Wen, Yongbin Liu, Chunping Ouyang, Qiang Lin, and Tong~Lee Chung.
\newblock Enhanced prototypical network for few-shot relation extraction.
\newblock {\em Inf. Process. Manag.}, 58(4):102596, April 2021.

\bibitem[\protect\citeauthoryear{Wu \bgroup \em et al.\egroup
  }{2021}]{wu2021universal}
Aming Wu, Yahong Han, Linchao Zhu, and Yi~Yang.
\newblock Universal-prototype enhancing for few-shot object detection.
\newblock In {\em Proceedings of the IEEE/CVF International Conference on
  Computer Vision}, pages 9567--9576, 2021.

\bibitem[\protect\citeauthoryear{Yang \bgroup \em et al.\egroup
  }{2021a}]{yang2022sega}
Fengyuan Yang, Ruiping Wang, and Xilin Chen.
\newblock {SEGA:} semantic guided attention on visual prototype for few-shot
  learning.
\newblock {\em CoRR}, abs/2111.04316, November 2021.

\bibitem[\protect\citeauthoryear{Yang \bgroup \em et al.\egroup
  }{2021b}]{yang2021entity}
Shan Yang, Yongfei Zhang, Guanglin Niu, Qinghua Zhao, and Shiliang Pu.
\newblock Entity concept-enhanced few-shot relation extraction.
\newblock In {\em ACL/IJCNLP}, pages 987--991, Virtual Event, August 2021.
  Association for Computational Linguistics.

\bibitem[\protect\citeauthoryear{Ye and Ling}{2019}]{ye2019multi}
Zhixiu Ye and Zhenhua Ling.
\newblock Multi-level matching and aggregation network for few-shot relation
  classification.
\newblock In {\em ACL}, pages 2872--2881, Florence, Italy, July--August 2019.
  Association for Computational Linguistics.

\bibitem[\protect\citeauthoryear{Zhang \bgroup \em et al.\egroup
  }{2020}]{zhang2020target}
Ji~Zhang, Chengyao Chen, Pengfei Liu, Chao He, and Cane~Wing{-}Ki Leung.
\newblock Target-guided structured attention network for target-dependent
  sentiment analysis.
\newblock {\em Trans. Assoc. Comput. Linguistics}, 8:172--182, 2020.

\end{thebibliography}
\end{document}